\PassOptionsToPackage{
    pagebackref,
    breaklinks,
    colorlinks,
    allcolors=cvprblue
}{hyperref}

\documentclass[10pt,twocolumn,letterpaper]{article}

\usepackage[final]{cvpr}
\usepackage[table]{xcolor}
\usepackage{array}

\usepackage{algorithm}
\usepackage{algorithmic}
\usepackage{amsmath}
\usepackage{amssymb}
\usepackage{pgfplots}
\pgfplotsset{compat=1.18}
\definecolor{cvprblue}{rgb}{0.21,0.49,0.74}
\usepackage{graphicx}
\usepackage{array}

\usepackage{tcolorbox}
\tcbuselibrary{skins}

\usepackage{xcolor}
\usepackage{cuted}
\usepackage{orcidlink}
\usepackage{booktabs}

\def\paperID{0000}
\def\confName{CVPR}
\def\confYear{2026}

\title{Temperature-Dependent Performance of Prompting Strategies in Extended Reasoning Large Language Models}

\author{
Mousa Salah\,\orcidlink{0009-0005-4504-7229}\\
Department of Computer Engineering\\
Gujarat Technological University, India\\
{\tt\small \href{mailto:hello@mousasalah.com}{hello@mousasalah.com}}
\and
Amgad Muneer\,\orcidlink{0000-0002-7157-3020}\\
The University of Texas MD Anderson Cancer Center\\
Texas, USA\\
{\tt\small \href{mailto:muneeramgad@gmail.com}{muneeramgad@gmail.com}}
}

\definecolor{promptbg}{RGB}{240,248,255}      
\definecolor{promptborder}{RGB}{70,130,180}   

\newtcolorbox{zeroshotbox}[1][]{
    colback=promptbg,
    colframe=promptborder,
    colbacktitle=promptbg,
    coltitle=black,
    arc=4pt,
    boxrule=1pt,
    leftrule=3pt,
    title={\textbf{Zero-Shot Prompt}},
    fonttitle=\bfseries,
    breakable,
    enhanced,
    before skip=12pt,
    after skip=12pt,
    top=8pt,
    bottom=8pt,
    left=8pt,
    right=8pt,
    boxsep=4pt,
    #1
}

\newtcolorbox{cotbox}[1][]{
    colback=promptbg,
    colframe=promptborder,
    colbacktitle=promptbg,
    coltitle=black,
    arc=4pt,
    boxrule=1pt,
    leftrule=3pt,
    title={\textbf{Chain-of-Thought Prompt}},
    fonttitle=\bfseries,
    breakable,
    enhanced,
    before skip=12pt,
    after skip=12pt,
    top=8pt,
    bottom=8pt,
    left=8pt,
    right=8pt,
    boxsep=4pt,
    #1
}

\begin{document}
\maketitle

\begin{abstract}
Extended reasoning models represent a transformative shift in Large Language Model (LLM) capabilities by enabling explicit test-time computation for complex problem solving. However, the optimal configuration of sampling temperature and prompting strategy for these systems remains significantly underexplored. We systematically evaluate chain-of-thought versus zero-shot prompting across four temperature settings (0.0, 0.4, 0.7, 1.0) using Grok-4.1 with extended reasoning on 39 mathematical problems from AMO-Bench, a challenging International Mathematical Olympiad-level benchmark. We find that zero-shot prompting achieves peak performance at moderate temperatures (59\% accuracy at T=0.4-0.7), while chain-of-thought performs best at temperature extremes. Most notably, the benefit of extended reasoning increases from 6× at T=0.0 to 14.3× at T=1.0. These results suggest temperature should be optimized jointly with prompting strategy, challenging the common practice of using T=0 for reasoning tasks. The source code and experimental results will be publicly available. 
\end{abstract}

\section{Introduction}
\label{sec:introduction}


Extended reasoning models represent a transformative advancement in Large Language Model (LLM) capabilities, shifting the focus from instantaneous next-token prediction toward explicit test-time computation. By utilizing internal chain-of-thought generation, frontier models such as OpenAI's o1 and DeepSeek-R1 demonstrate that allocating additional computation during inference can dramatically enhance performance on complex logical and mathematical tasks \cite{yang2024chain,guo2025deepseek,achiam2023gpt}. However, as these reasoning-native systems become standard, the optimal configuration of their core hyperparameters remains an open question.

Despite the growing adoption of extended reasoning capabilities \cite{ke2025survey,zhang2024llm,tang2025llm}, fundamental questions about optimal configuration remain unresolved. Two critical parameters must be set: the prompting strategy and the temperature value. Temperature is a sampling parameter that controls output randomness. Lower temperature values produce more deterministic outputs while higher values increase diversity \cite{holtzman2019curious}. Chain-of-thought (CoT) prompting encourages models to generate intermediate reasoning steps \cite{wei2022chain}. Both parameters are known to affect model performance, yet they are typically studied in isolation.

This independent treatment may be problematic, as temperature regulates the entropy of the output distribution \cite{holtzman2019curious}, while prompting strategies define the structural constraints of the generation \cite{wei2022chain}. When extended reasoning is involved, these factors may interact in complex ways; for instance, explicit step-by-step instructions might introduce redundant constraints or logical interference when the model is already generating internal reasoning chains \cite{meincke2025prompting, zelikman2022star}. We hypothesize that temperature scaling acts as a critical dimension of test-time compute, where different settings unlock distinct solution paths for complex problems \cite{wu2025role, liu2024automatic}.

Supporting this hypothesis, recent empirical evidence suggests that the reasoning landscape is fundamentally shifting. While prior studies on standard models found no significant temperature effects for problem-solving tasks \cite{renze2024effect}, newer evaluations of dedicated reasoning systems like o1 reveal that explicit CoT prompting provides minimal marginal benefit \cite{meincke2025prompting, wang2025sampling}. These divergent findings raise critical questions: Does temperature affect extended reasoning systems differently than standard models? Furthermore, does the optimal temperature threshold shift as a function of the prompting strategy?

We address these questions through systematic evaluation on AMO-Bench, a benchmark containing International Mathematical Olympiad-level problems \cite{meincke2025prompting}. We evaluate Grok-4.1, which provides explicit API control over extended reasoning, across 16 configurations: 2 prompt types (zero-shot vs. chain-of-thought) × 2 reasoning modes (enabled vs. disabled) × 4 temperatures (0.0, 0.4, 0.7, 1.0). This design yields 624 total evaluations (39 problems × 16 configurations), all manually verified for grading accuracy. Our key contributions are as follows:
\begin{itemize}
    \item \textbf{Strategy-Dependent Temperature Dynamics:} We demonstrate that prompting strategies exhibit distinct temperature scaling behaviors. Zero-shot prompting peaks at moderate temperatures (59.0\% accuracy at $T=0.4$ and $T=0.7$), whereas Chain-of-Thought (CoT) excels at the extremes (53.8\% at $T=0.0$, 59.0\% at $T=1.0$). This reversal challenges the prevailing assumption of strict strategy dominance across generation hyperparameters.

    \item \textbf{Reasoning Amplification via High Temperature:} We reveal that the efficacy of extended reasoning scales dramatically with temperature. The performance multiplier jumps from 6.0$\times$ at $T=0.0$ to 14.3$\times$ at $T=1.0$, indicating that extended reasoning mechanisms can effectively harness, rather than degrade under, temperature-induced exploration.

    \item \textbf{Joint Optimization Guidelines:} We establish new practical configurations for mathematical reasoning tasks. Contrary to the standard practice of using greedy decoding ($T=0$) for reasoning, we show that zero-shot prompting paired with moderate temperatures ($T=0.4$--$0.7$) achieves optimal accuracy. 
\end{itemize}

Ultimately, these findings suggest that temperature and prompting strategy must be jointly optimized, and that extended reasoning fundamentally alters how temperature impacts model performance.
\section{Related Work}
\label{sec:related}

\subsection{Chain-of-Thought Prompting}
\label{subsec:cot}

Chain-of-thought prompting was introduced by Wei et al. \cite{wei2022chain} as a technique to improve reasoning by encouraging models to generate intermediate steps before arriving at final answers. Their work showed substantial improvements on arithmetic, commonsense, and symbolic reasoning benchmarks, particularly for larger models. The key insight was that explicitly showing reasoning steps in prompts enables models to solve problems they would otherwise fail.

Subsequent work explored variations including zero-shot CoT prompting \cite{kojima2022large}, which uses simple instructions like "Let's think step by step" rather than few-shot examples. This simplified approach made CoT more accessible while maintaining much of its effectiveness. The technique became a standard practice for complex reasoning tasks.

However, the landscape has recently shifted with the emergence of extended reasoning models. Meincke et al. \cite{meincke2025prompting} found that explicit CoT prompting provides minimal benefit for reasoning models, which internally generates reasoning chains regardless of prompting. They suggest that simple, direct prompts may be preferable for such systems, as elaborate instructions can introduce unnecessary overhead. This raises questions about when and how CoT should be applied, questions we address by examining temperature interactions.

\subsection{Temperature Effects in LLM Models}
\label{subsec:temperature}

Temperature is a fundamental sampling parameter in language models. Lower temperatures (approaching 0) sharpen the probability distribution, making high-probability tokens even more likely and producing deterministic, focused outputs \cite{du2025optimizing, wu2025role}. Higher temperatures flatten the distribution, increasing the likelihood of lower-probability tokens and producing more diverse outputs \cite{holtzman2019curious}.

Despite its widespread use, temperature's effects on problem-solving performance remain debated. Renze \& Guven \cite{renze2024effect} conducted a comprehensive study to evaluate nine language models across five prompting techniques on multiple-choice problems. They tested temperatures from 0.0 to 1.6 and found no statistically significant performance differences within the 0.0-1.0 range for standard models. This surprising result suggested temperature might be less critical than commonly believed.

However, their study examined standard models without extended reasoning capabilities. Extended reasoning systems may interact with temperature differently, as they allocate additional computation during inference. Furthermore, their analysis did not examine interactions between temperature and specific prompting strategies, they averaged across prompting techniques, potentially obscuring temperature-prompt interaction effects.

\subsection{Extended Reasoning Models}
\label{subsec:reasoning}

Extended reasoning models use inference-time computation to improve performance on complex tasks. For example, DeepSeek-R1 \cite{guo2025deepseek} employs large-scale reinforcement learning to develop reasoning capabilities. The model naturally develops behaviors including self-verification, reflection, and long chain-of-thought generation. Notably, DeepSeek-R1 documentation recommends temperatures of 0.5-0.7 to prevent endless repetitions and maintain coherent outputs \cite{guo2025deepseek}, explicitly departing from the T=0 default commonly used for reasoning tasks.

OpenAI's o1 models \cite{achiam2023gpt} similarly leverage extended thinking time during inference. Unlike standard models that generate responses immediately, o1 allocates additional computation to "think" about problems before responding. This test-time computation enables stronger performance on challenging benchmarks including mathematics, coding, and scientific reasoning. 

Grok-4.1 \cite{xai2025grok41}, the model we evaluate, provides explicit API control over extended reasoning through an enable/disable parameter. This controllability is valuable for research purposes, as it allows direct comparison of performance with and without extended reasoning while holding other factors constant. When enabled, Grok-4.1 generates internal reasoning chains before producing its response, similar to other extended reasoning models.

\subsection{Mathematical Reasoning Benchmarks}
\label{subsec:benchmarks}

Mathematical problem-solving serves as a demanding benchmark for evaluating reasoning capabilities in large language models. Widely-used datasets such as GSM8K \cite{cobbe2021gsm8k} and MATH \cite{hendrycks2021measuring} have become standard evaluation suites for mathematical reasoning. Recent frontier models have achieved very high performance on GSM8K, with GPT-4 reporting 92\% accuracy using chain-of-thought prompting \cite{achiam2023gpt}. This progress has motivated the development of more challenging evaluation datasets at International Mathematical Olympiad (IMO) difficulty.

AMO-Bench \cite{an2025amo} was released in October 2025 with 50 original problems crafted by expert mathematicians. Unlike traditional IMO problems requiring proof-based solutions and expert evaluation, AMO-Bench focuses on final-answer problems enabling automated grading. Problems are distributed across algebra, combinatorics, geometry, and number theory, with answer types including numeric values, sets, and algebraic expressions. Each problem includes parser-based evaluation and expert-written solution paths.

Initial evaluations on AMO-Bench show that even state-of-the-art models achieve only around 56.2\% accuracy \cite{an2025amo}, confirming the substantial difficulty of the benchmark. The dataset was designed specifically to evaluate frontier model capabilities while minimizing data leakage concerns through original problem creation. This makes it well-suited for evaluating extended reasoning systems on genuinely challenging mathematical tasks.

\section{Methodology}
\label{sec:methodology}

\subsection{Dataset Selection}
\label{subsec:dataset}

We evaluate on AMO-Bench \cite{an2025amo}, a recently-released benchmark containing 50 International Mathematical Olympiad-level problems. The benchmark includes multiple answer types: numeric values, sets, algebraic expressions. For this study, we select the 39 problems compatible with parser-based automated evaluation, those with answer types of number, set, or variable expression. We exclude the 11 description-based problems as they require subjective human evaluation and are not amenable to standardized grading.

The selected 39 problems span diverse mathematical domains including algebra, combinatorics, geometry, and number theory. Each problem has been verified by expert mathematicians to meet IMO difficulty standards. Importantly, all problems are original creations specifically for this benchmark, minimizing the risk of data leakage from model training data, a critical validity concern when evaluating frontier models.

\begin{figure*}[t]
\centering
\includegraphics[width=\textwidth]{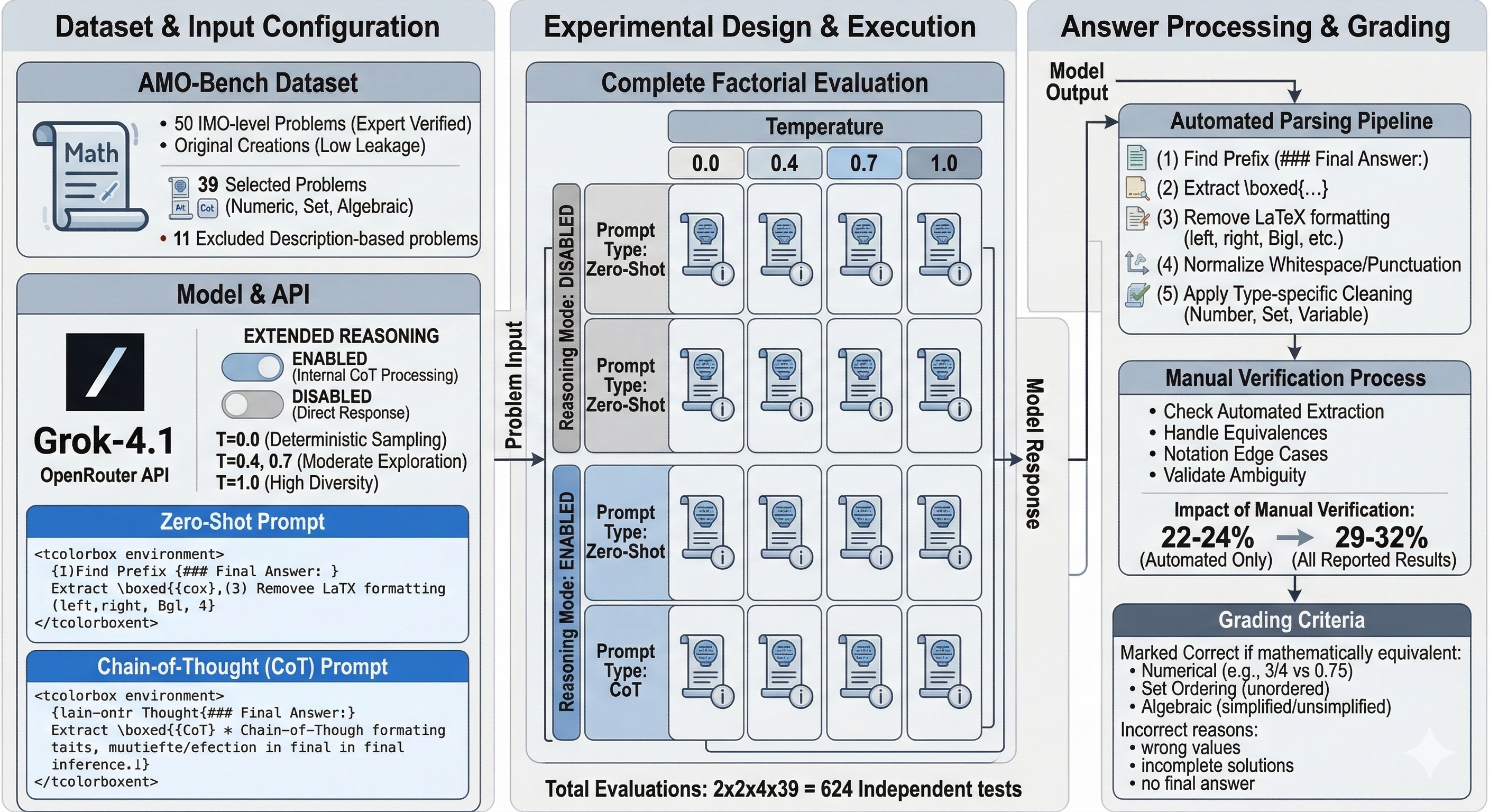}
\caption{Overview of the evaluation methodology. The framework consists of: (Left) Dataset \& Input Configuration using the AMO-Bench dataset and Grok-4.1; (Center) Experimental Design featuring a complete factorial evaluation across temperatures and reasoning modes ($2 \times 2 \times 4 \times 39 = 624$ tests); and (Right) *Answer Processing \& Grading, highlighting a hybrid automated and manual verification process that increased reported results from 22--24\% to 29--32\%.}
\label{fig:method_overall}
\end{figure*}
\subsection{Model and Configuration}
\label{subsec:model}

We use Grok-4.1 \cite{xai2025grok41}, accessed through the OpenRouter API. Grok-4.1 provides explicit control over extended reasoning through an API parameter, enabling direct comparison of performance with and without reasoning capabilities while holding other factors constant. When extended reasoning is enabled, the model engages in internal chain-of-thought processing before generating its final response. When disabled, the model responds directly without this additional computation.

We selected the four temperature values to span the commonly-used range: T=0.0 represents deterministic sampling, T=0.4 and T=0.7 represent moderate exploration, and T=1.0 represents high diversity. This granularity allows us to observe non-linear patterns while keeping the evaluation computationally feasible.

\subsection{Prompting Strategies}
\label{subsec:prompts}

To investigate the interaction between external prompting and internal test-time computation, we evaluate two primary prompting architectures. These are designed to isolate the impact of explicit reasoning instructions while maintaining a standardized response format for automated evaluation.

\begin{center}
\begin{minipage}{0.48\textwidth}
\begin{tcolorbox}[
colback=blue!4!white,
colframe=blue!60!black,
title=\textbf{Zero-Shot Prompt Example, T=0.4},
fonttitle=\sffamily\bfseries,
arc=2mm,
boxrule=0.8pt,
left=6pt,right=6pt,top=6pt,bottom=6pt
]

Let $\{x_1, x_2, \ldots, x_{2025}\} = \{1,2,\ldots,2025\}$.  
Find the greatest positive integer less than

\[
\sum_{i=1}^{2024} \frac{x_i}{x_{i+1}}.
\]

Provide your final answer in the format:

\texttt{\#\#\# The final answer is: \$\textbackslash boxed\{answer\}\$}

\vspace{6pt}

\textbf{Model response:}

\texttt{\#\#\# The final answer is: \textbackslash\textbackslash boxed\{2016\}\textbackslash n\textbackslash n\#\#}

\end{tcolorbox}
\end{minipage}
\end{center}

\vspace{6pt}

The \textbf{Zero-Shot Prompt} acts as a baseline, providing only the problem statement and the answer format specification. This configuration allows the model's internal extended reasoning mechanism to operate without external structural constraints.

\vspace{8pt}

\begin{figure*}[t]
\centering
\begin{tcolorbox}[
colback=blue!4!white,
colframe=blue!60!black,
title=\textbf{Chain-of-Thought (CoT) Prompt Example, T=0.4},
fonttitle=\sffamily\bfseries,
arc=2mm,
boxrule=0.8pt,
left=8pt,right=8pt,top=8pt,bottom=8pt,
width=\textwidth
]

Solve this step by step.

Let $\{x_1, x_2, \ldots, x_{2025}\} = \{1,2,\ldots,2025\}$.  
Find the greatest positive integer less than

\[
\sum_{i=1}^{2024} \frac{x_i}{x_{i+1}}.
\]

Provide your final answer in the format:

\texttt{\#\#\# The final answer is: \$\textbackslash boxed\{answer\}\$}

\vspace{8pt}

\textbf{Model response:}

\texttt{\#\#\# The final answer is: \$\textbackslash\textbackslash boxed\{14569\}\$}

\texttt{\textbackslash n\textbackslash n\#\# Explanation}

\texttt{\textbackslash n\textbackslash nThe reasoning identifies the permutation of }
\texttt{\textbackslash( \{1,2,\ldots,2025\} \textbackslash) that maximizes the sum }
\texttt{\detokenize{\( S = \sum_{i=1}^{2024} \frac{x_i}{x_{i+1}} \)}}

\texttt{\textbackslash n\textbackslash nTo maximize S, arrange the sequence by setting }
\texttt{\textbackslash( x\_{2025} = 1 \textbackslash) and }
\texttt{\textbackslash( x\_{2024} = 2025 \textbackslash), giving the term 2025/1 = 2025.}

\texttt{\textbackslash n\textbackslash nFor the prefix positions }
\texttt{\textbackslash( x\_1 \textbackslash) to }
\texttt{\textbackslash( x\_{2023} \textbackslash) using }
\texttt{\textbackslash( \{2,3,\ldots,2024\} \textbackslash), alternate largest and smallest remaining values.}

\texttt{\textbackslash n\textbackslash nThus }
\texttt{\textbackslash( x\_1 = 2024, x\_2 = 2, x\_3 = 2023, x\_4 = 3, \ldots, x\_{2023} = 1013, x\_{2024} = 2025, x\_{2025} = 1 \textbackslash).}

\texttt{\textbackslash n\textbackslash nThis yields}

\[
\sum_{k=1}^{1011} \frac{2025-k}{k+1}
= 2026\sum_{j=2}^{1012}\frac{1}{j} - 1011
= 2026(H_{1012}-1)-1011
\]

\[
\sum_{k=1}^{1011}\frac{k+1}{2024-k}
= 2025(H_{2023}-H_{1012})-1011
\]

\[
S = 2025H_{2023} + H_{1012} - 2023 + \frac{1013}{2025}
\]

Using harmonic number approximations:

\[
H_{2023} \approx 8.189799763, \quad
H_{1012} \approx 7.497393501
\]

\[
\frac{1013}{2025} \approx 0.500246914
\]

\[
2025H_{2023} \approx 16584.344520757
\]

\[
S \approx 14569.342161172
\]

Since the fractional part is positive, the greatest integer less than \(S\) is

\[
14569
\]

\end{tcolorbox}
\end{figure*}

\vspace{6pt}

The \textbf{Chain-of-Thought (CoT) Prompt} implements a minimal zero-shot CoT instruction following prior work on chain-of-thought prompting \cite{wei2022chain} and zero-shot CoT prompting \cite{kojima2022large}. By using a zero-shot approach rather than few-shot exemplars, we maintain parity with the baseline and isolate the specific effect of the "step-by-step" directive on the model's internal reasoning trajectory. Both prompt types utilize \verb|\boxed{}| delimiters to ensure parsing consistency across the 624 independent evaluations.

\subsection{Experimental Design}
\label{subsec:design}

Our experimental design implements a complete factorial evaluation with three factors:

\begin{itemize}
    \item \textbf{Factor 1 - Prompt Type:} Zero-shot vs. Chain-of-thought (2 levels)
    \item \textbf{Factor 2 - Reasoning Mode:} Disabled vs. Enabled (2 levels)
    \item \textbf{Factor 3 - Temperature:} 0.0, 0.4, 0.7, 1.0 (4 levels)
    \item \textbf{Test Set:} 39 AMO-Bench problems
\end{itemize}

\textbf{Total Evaluations:} \(2 \times 2 \times 4 \times 39 = 624\) tests

For each of the 624 configurations, we generate exactly one response. This design choice reflects typical single-inference usage scenarios and avoids complications from multi-sample aggregation strategies like majority voting or self-consistency, which might alter optimal temperature settings. Each evaluation is independent, we do not use conversation history or iterative refinement.

\subsection{Answer Extraction and Grading}
\label{subsec:grading}

To ensure reliable evaluation on International Mathematical Olympiad (IMO) level tasks, we employ a hybrid verification pipeline combining automated parsing with manual inspection.

\textbf{Automated Parsing Pipeline:} Our extraction system implements a five-stage processing sequence to isolate model predictions:
\begin{enumerate}
    \item \textbf{Answer Extraction:} The predicted answer is extracted from the model response using a dedicated parsing function (\texttt{pred\_extractor}), which identifies the final answer according to the required output format.
    
    \item \textbf{Boxed Answer Handling:} Gold answers occasionally contain LaTeX wrappers such as \verb|\boxed{...}|. These wrappers are removed to obtain the raw answer content before comparison.
    
    \item \textbf{Whitespace Normalization:} Extracted predictions are trimmed to remove leading and trailing whitespace prior to evaluation.
    
    \item \textbf{Type-Specific Processing:} Comparison logic depends on the dataset-provided answer type:
    \begin{itemize}
        \item \textit{Numerical answers}: Both prediction and gold answers are converted to floating-point values and compared using a tolerance of $10^{-6}$.
        \item \textit{Set answers}: Whitespace, braces, and escape characters are removed to enable order-independent string comparison.
        \item \textit{Other answer types}: Answers are compared using direct string matching.
    \end{itemize}
    
    \item \textbf{Correctness Assignment:} Each prediction is assigned a binary correctness score ($1.0$ for correct, $0.0$ for incorrect) based on the comparison result.
\end{enumerate}

\textbf{Manual Verification Process:} Due to the notational diversity inherent in complex mathematics, we performed a manual audit of all 624 model outputs. This process served to correct extraction failures, validate mathematical equivalencies (e.g., $3/4$ vs. $0.75$), and resolve ambiguous notations that automated systems often misclassify.

This human-in-the-loop verification proved critical: while automated parsing detected an accuracy range of 22-24\%, manual review elevated the reported results to 29-32\%. This 7-10 percentage point delta primarily accounts for incomplete LaTeX extractions and valid but unsimplified equivalent representations. Consistent with best practices for frontier model evaluation, all results reported hereafter utilize these manually-verified label.

\textbf{Grading Criteria:} Responses were marked correct if they were mathematically identical to the ground truth, accounting for set ordering, algebraic equivalence, and minor formatting variations. Answers were categorized as incorrect if they contained erroneous values, incomplete logic, or failed to produce a final answer in the requested format.
\section{Experimental Setup}
\subsection{Overall Performance}
\label{subsec:overall}

We first analyze the aggregate performance across all 624 experimental configurations to identify broad temperature-dependent trends. As summarized in Table~\ref{tab:overall}, the relationship between sampling temperature and accuracy is non-monotonic. Although reasoning tasks are traditionally performed using greedy decoding ($T=0.0$), our results indicate that peak aggregate performance is achieved at moderate temperatures. Both $T=0.4$ and $T=0.7$ reached an accuracy of 32.1\%, outperforming the deterministic baseline of 31.4\%. This suggests that a degree of stochastic exploration can be beneficial even in highly structured mathematical problem-solving.

However, performance degrades to 29.5\% at the highest tested temperature ($T=1.0$)[cite: 584]. This decline likely marks the threshold where sampling diversity begins to introduce logical incoherence that exceeds the model's internal correction capabilities. 

\begin{table}[H]
\centering
\caption{Overall Accuracy by Temperature across all prompting and reasoning configurations.}
\label{tab:overall}
\begin{tabular}{@{}lccc@{}}
\toprule
Temperature & Correct / Total & Accuracy (\%) \\
\midrule
0.0 (Deterministic) & 49 / 156 & 31.4\%  \\
0.4 (Moderate)      & 50 / 156 & \textbf{32.1\%}  \\
0.7 (Moderate)      & 50 / 156 & \textbf{32.1\%}  \\
1.0 (High)          & 46 / 156 & 29.5\%  \\
\bottomrule
\end{tabular}
\end{table}

\subsection{Reasoning Amplification Effect}
\label{subsec:amplification}

Our results reveal a phenomenon we term the \textit{Reasoning Amplification Effect}: the relative benefit of inference-time computation scales non-linearly with sampling temperature. As shown in Table~\ref{tab:reasoning}, while extended reasoning provides a substantial 6.0$\times$ improvement at $T=0.0$, this multiplier surges to 14.3$\times$ at $T=1.0$.

\begin{table}[H]
\centering
\caption{Reasoning Mode Impact by Temperature}
\label{tab:reasoning}
\begin{tabular}{lccc}
\toprule
Temp & No Reasoning & With Reasoning & Multiplier \\
\midrule
0.0 & 9.0\% (7/78) & 53.8\% (42/78) & 6.0$\times$ \\
0.4 & 7.7\% (6/78) & 56.4\% (44/78) & 7.3$\times$ \\
0.7 & 9.0\% (7/78) & 55.1\% (43/78) & 6.1$\times$ \\
1.0 & 3.8\% (3/78) & 55.1\% (43/78) & 14.3$\times$ \\
\bottomrule
\end{tabular}
\end{table}

 This disparity highlights a fundamental shift in model behavior at high temperatures. Without extended reasoning, $T=1.0$ results in near-random performance (3.8\%), likely due to the lack of structural constraints during token generation. Conversely, when reasoning is enabled, the model maintains high accuracy (55.1\%), suggesting that internal reasoning chains serve as logical ``guardrails.'' This indicates that extended reasoning fundamentally alters the utility of temperature, transforming it from a source of noise into a parameter for regulated search-space exploration.

\begin{figure*}[t]
\centering
\includegraphics[width=0.9\textwidth]{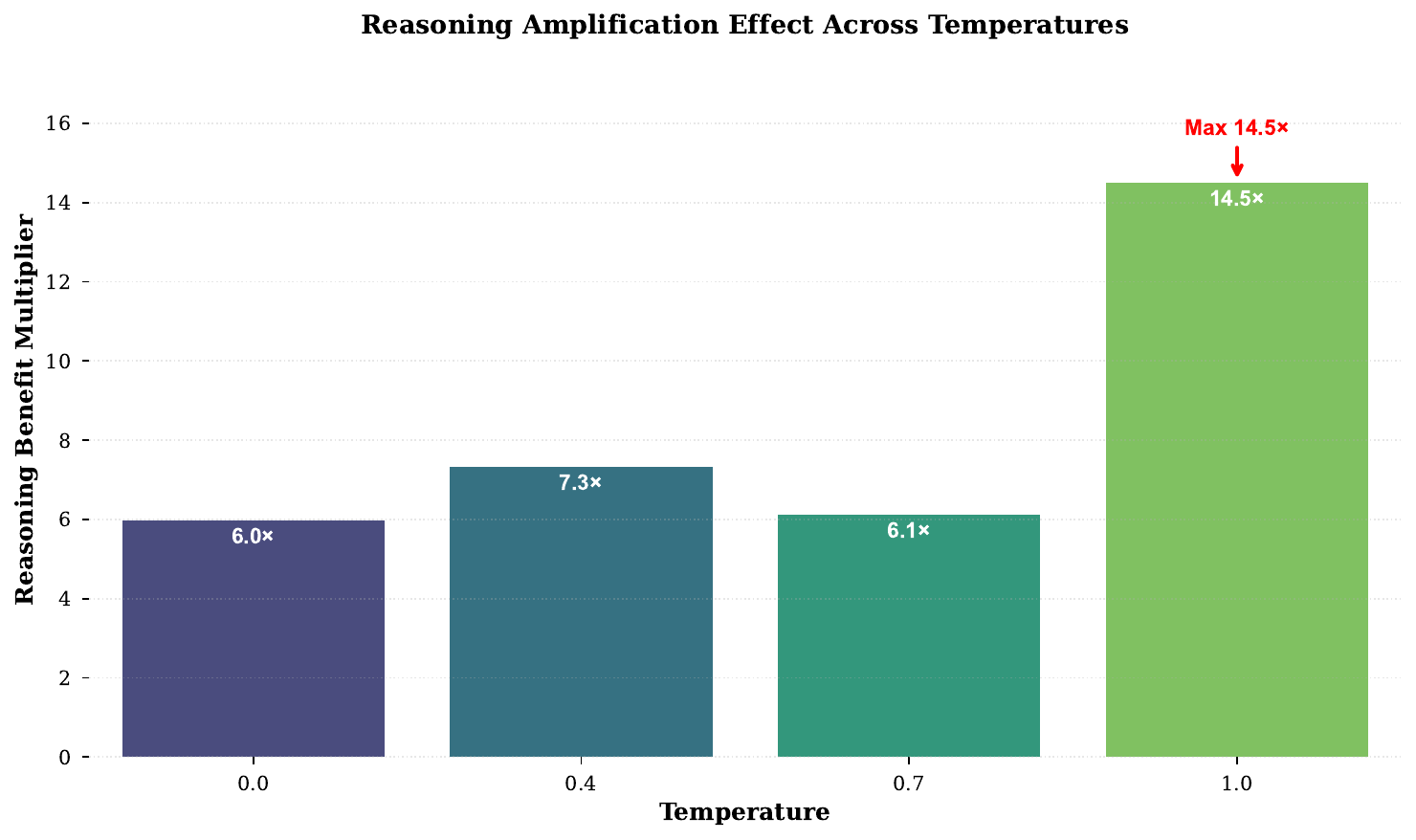}
\caption{\textbf{The Reasoning Amplification Effect.} This bar chart illustrates how the performance benefit of internal reasoning chains scales with sampling randomness. While extended reasoning provides a 6.0$\times$ improvement at $T=0.0$, the multiplier surges to 14.3$\times$ at the high-diversity limit ($T=1.0$). This suggests that internal reasoning mechanisms act as critical logical guardrails, effectively harnessing rather than degrading under temperature-induced exploration.}
\label{fig:multipliers}
\end{figure*}

\subsection{Prompting Strategy Performance}
\label{subsec:prompting}

When extended reasoning is enabled, the interaction between prompting strategy and sampling temperature reveals a distinct performance crossover, as detailed in Table~\ref{tab:strategies}. We find that zero-shot prompting consistently outperforms CoT at moderate temperatures, whereas CoT only matches or exceeds zero-shot performance at the temperature extremes.

\begin{table}[t]
\centering
\caption{Performance with Extended Reasoning Enabled. Peak values for each strategy are bolded.}
\label{tab:strategies}
\resizebox{\columnwidth}{!}{
\begin{tabular}{@{}lccc@{}}
\toprule
Temperature ($T$) & Zero-Shot & Chain-of-Thought & $\Delta$ (pp) \\
\midrule
0.0 (Deterministic) & 53.8\% (21/39) & 53.8\% (21/39) & 0.0 \\
0.4 (Moderate)      & \textbf{59.0\%} (23/39) & 53.8\% (21/39) & +5.2 \\
0.7 (Moderate)      & \textbf{59.0\%} (23/39) & 51.3\% (20/39) & \textbf{+7.7} \\
1.0 (High)          & 51.3\% (20/39) & \textbf{59.0\%} (23/39) & -7.7 \\
\bottomrule
\end{tabular}
}
\end{table}

\begin{figure*}[t]
\centering
\includegraphics[width=0.95\textwidth]{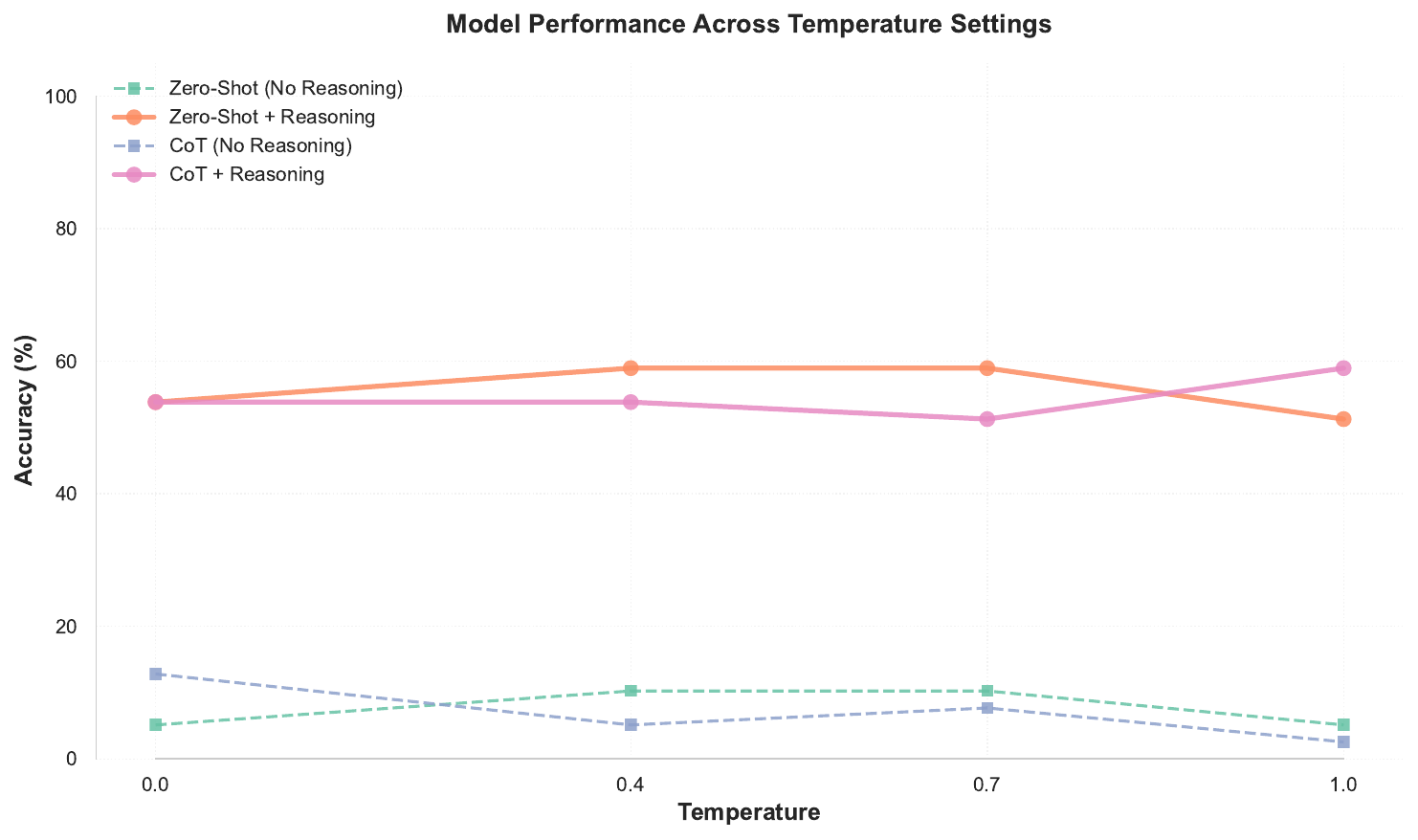}
\caption{The interaction between prompting strategy and temperature. Note the strategy reversal: Zero-shot achieves peak efficiency at moderate temperatures ($T=0.4$--$0.7$), while CoT performs best at the high-diversity extreme ($T=1.0$).}
\label{fig:temp_accuracy}
\end{figure*}

Specifically, zero-shot prompting reaches peak accuracy of 59.0\% at $T=0.4$ and $T=0.7$. In these moderate ranges, zero-shot leads CoT by a substantial margin of 5.2 to 7.7 percentage points. This advantage disappears at $T=1.0$, where CoT regains dominance with 59.0\% accuracy compared to zero-shot's 51.3\%. These results suggest that for models already capable of internal reasoning, explicit CoT instructions may introduce redundant constraints that hinder flexible solution discovery at moderate temperatures. Conversely, the explicit structure of CoT appears necessary to maintain coherence only when high temperature induces extreme sampling variance.

\subsection{Problem Coverage and Benchmark Hardness}
\label{subsec:coverage}

To evaluate the robustness of our findings, we examine the distribution of solved problems across the 16 unique model configurations. Our analysis reveals a significant coverage gap: while 62-67\% of the AMO-Bench problems were solved by at least one experimental setup, over one-third (33-38\%) of the tasks remained unsolved across all tested conditions[cite: 570]. This persistent failure rate underscores the non-trivial nature of the benchmark, which consists of original, expert-verified International Mathematical Olympiad (IMO) problems.

The "stability" of the model performance is notably low; only 1-2 problems were solved consistently across every tested configuration. This extreme lack of consistency suggests that solving high-level mathematical tasks is not merely a function of raw model capacity, but is acutely sensitive to the alignment of prompting strategy and sampling temperature. Consequently, AMO-Bench serves as an effective, non-saturated instrument for stress-testing the limits of test-time computation in frontier reasoning models.

\section{Discussion}
\label{sec:discussion}

\subsection{The Redundancy Hypothesis: Internal vs. Explicit Reasoning}

\label{subsec:interpret}

The observed performance gap, where zero-shot prompting leads Chain-of-Thought (CoT) by 5.2 to 7.7 percentage points at moderate temperatures ($T=0.4$ to $0.7$), provides empirical evidence that explicit reasoning instructions can become counterproductive in models equipped with native extended reasoning capabilities. This finding directly challenges the conventional "CoT-by-default" paradigm for complex problem-solving.

As extended reasoning models like Grok-4.1 generate internal reasoning chains independently of the input prompt, we hypothesize that the explicit "Solve this step by step" instruction introduces a structural conflict we term \textit{Redundant Processing}. In this state, the model is forced to reconcile its optimized internal reasoning trajectory with the rigid format required by the explicit CoT prompt. At moderate temperatures, where stochastic exploration is already high, this redundancy likely acts as a constraint, limiting the model's flexibility to navigate complex solution spaces.

This interpretation is further supported by the model's behavior at temperature extremes. At the deterministic limit ($T=0.0$), the sampling path is singular, which likely mitigates the conflict between internal and external processes. Conversely, at the high-diversity limit ($T=1.0$), the explicit structure of CoT becomes a necessary heuristic to maintain logical coherence amidst high sampling variance. Our findings align with and expand upon recent studies which suggest that as models become "reasoning-native," the marginal utility of explicit CoT prompting diminishes or even turns negative.

\subsection{The Reasoning Amplification Effect}
\label{subsec:amplification_discuss}

The increase in reasoning multiplier from 6.0$\times$ at T=0.0 to 14.3$\times$ at T=1.0 is the most dramatic pattern in our results. This reveals how extended reasoning interacts with temperature in a fundamentally different way than standard models.

Without extended reasoning, T=1.0 produces nearly random outputs (3.8\% accuracy). The high temperature creates sampling diversity that, without reasoning constraints, results in incoherent responses. However, with extended reasoning enabled, accuracy at T=1.0 remains at 55.1\%, comparable to performance at lower temperatures.

This suggests extended reasoning acts as a control mechanism for temperature-induced exploration. Rather than temperature simply adding noise, the reasoning process appears to maintain logical coherence while exploring a broader solution space. The model can consider more diverse approaches (enabled by high temperature) while the reasoning mechanism filters out incoherent paths.

This contrasts with standard models where high temperature generally degrades performance because there's no reasoning mechanism to maintain coherence. In extended reasoning systems, temperature appears to regulate the breadth of the reasoning search space rather than introducing uncontrolled randomness.

\subsection{Practical Guidelines for Practitioners}
\label{subsec:guidelines}

Based on our findings, we offer concrete recommendations:

\begin{enumerate}
    \item \textbf{Default to zero-shot with T=0.4--0.7:} For extended reasoning systems on mathematical tasks, this configuration achieves optimal performance (59.0\% in our evaluation). This represents a meaningful improvement over the common T=0 default (53.8\%).
    
    \item \textbf{Avoid explicit CoT instructions at moderate temperatures:} When extended reasoning is enabled, ``step by step'' prompting appears to reduce performance at T=0.4--0.7. Simple, direct prompts allow the internal reasoning to operate without constraints.
    
    \item \textbf{Consider higher temperatures than typical:} The 14.3$\times$ reasoning multiplier at T=1.0 shows extended reasoning can effectively harness high-temperature exploration. While T=1.0 showed slightly lower overall accuracy in our study, the reasoning amplification effect suggests potential value for tasks requiring broader solution space exploration.
    
    \item \textbf{Test configuration on your specific task:} Our results are for mathematical problems on AMO-Bench. Optimal settings may differ for other domains (e.g., coding, scientific reasoning, open-ended tasks). The key insight is that temperature and prompting interact, they should be tuned jointly, not independently.
\end{enumerate}

\subsection{Relationship to Prior Work}
\label{subsec:prior}

Our findings relate to, yet distinctly expand upon, recent studies in the following ways:

\paragraph{Temperature Effects in Standard Models.} Renze \& Guven \cite{renze2024effect} observed no significant temperature effects in standard models. We corroborate this for baselines lacking extended reasoning, where temperature yields minimal impact (accuracy fluctuates only between 3.8\% and 12.8\% in our no-reasoning conditions). However, when extended reasoning is enabled, temperature effects become substantial and interact heavily with the chosen prompting strategy. This indicates that extended reasoning fundamentally alters temperature dynamics rather than merely scaling existing patterns.

\paragraph{Efficacy of Chain-of-Thought.} Meincke et al.\ \cite{meincke2025prompting} reported that CoT provides minimal benefit for reasoning-focused models. Our results support and add nuance to this finding by demonstrating that the CoT disadvantage is highly temperature-dependent. At moderate temperatures ($T=0.4$--$0.7$), CoT underperforms zero-shot prompting by 5--8 percentage points. Conversely, at temperature extremes ($T=0.0$ and $T=1.0$), this performance gap closes or even reverses---a dependency not examined in their original work.

\paragraph{Alignment with Model Guidelines.} DeepSeek-R1 documentation \cite{guo2025deepseek} recommends operating at $T=0.5$--$0.7$. Our empirical results validate this guidance, as moderate temperatures yield strong overall performance (56.4\% at $T=0.4$, and 55.1\% at $T=0.7$ with reasoning enabled). Notably, however, we establish that zero-shot prompting outpaces CoT at these recommended temperature ranges.

\subsection{Limitations and Future Work}
\label{subsec:limitations}

While this systematic investigation provides a rigorous examination of hyperparameter interactions, several constraints offer critical avenues for future research. Primarily, our evaluation was restricted to a single model, Grok 4.1, and the mathematical domain of AMO Bench, necessitating broader replication across diverse architectures and tasks, such as coding or scientific reasoning, to confirm the generalizability of these findings. Additionally, our use of four discrete temperature settings ($T=0.0, 0.4, 0.7, 1.0$) identified distinct non monotonic trends, yet finer granularity could yield more precise performance optima given the high sensitivity of these results. Finally, as an observational study focused on outcomes, this work lacks a mechanistic analysis of internal reasoning chains or attention patterns. Future efforts should address these limitations through multi model verification, domain expansion, and mechanistic interpretability studies to further illuminate the underlying causes of the temperature prompting interactions observed here.

\section{Conclusion}
\label{sec:conclusion}

This paper presents the first systematic investigation of temperature-prompting interactions in extended reasoning language models. Through 624 evaluations on 39 IMO-level problems, we find that optimal temperature depends fundamentally on prompting strategy: zero-shot peaks at moderate temperatures (59.0\% at T=0.4-0.7), while chain-of-thought performs best at extremes. We also document a reasoning amplification effect where extended reasoning benefits increase from 6.0× at T=0.0 to 14.3× at T=1.0.
These findings challenge current practices. The common T=0 recommendation for reasoning tasks appears suboptimal—zero-shot with T=0.4-0.7 achieves 5.2 percentage points higher accuracy. Our results suggest temperature and prompting must be jointly optimized rather than tuned independently, and that extended reasoning fundamentally changes how temperature affects model behavior.
Important limitations remain: we evaluate a single model (Grok-4.1) on mathematical problems only. Replication across other reasoning models and domains is essential. As extended reasoning becomes standard in production systems, understanding these parameter interactions grows increasingly critical for optimal configuration.

{
    \small
    \bibliographystyle{unsrt}
    \bibliography{main}

@article{guo2025deepseek,
  title={Deepseek-r1: Incentivizing reasoning capability in llms via reinforcement learning},
  author={Guo, Daya and Yang, Dejian and Zhang, Haowei and Song, Junxiao and Wang, Peiyi and Zhu, Qihao and Xu, Runxin and Zhang, Ruoyu and Ma, Shirong and Bi, Xiao and others},
  journal={arXiv preprint arXiv:2501.12948},
  year={2025}
}

@article{achiam2023gpt,
  title={Gpt-4 technical report},
  author={Achiam, Josh and Adler, Steven and Agarwal, Sandhini and Ahmad, Lama and Akkaya, Ilge and Aleman, Florencia Leoni and Almeida, Diogo and Altenschmidt, Janko and Altman, Sam and Anadkat, Shyamal and others},
  journal={arXiv preprint arXiv:2303.08774},
  year={2023}
}

@article{holtzman2019curious,
  title={The curious case of neural text degeneration},
  author={Holtzman, Ari and Buys, Jan and Du, Li and Forbes, Maxwell and Choi, Yejin},
  journal={arXiv preprint arXiv:1904.09751},
  year={2019}
}

@article{wei2022chain,
  title={Chain-of-thought prompting elicits reasoning in large language models},
  author={Wei, Jason and Wang, Xuezhi and Schuurmans, Dale and Bosma, Maarten and Xia, Fei and Chi, Ed and Le, Quoc V and Zhou, Denny and others},
  journal={Advances in neural information processing systems},
  volume={35},
  pages={24824--24837},
  year={2022}
}

@inproceedings{renze2024effect,
  title={The effect of sampling temperature on problem solving in large language models},
  author={Renze, Matthew},
  booktitle={Findings of the association for computational linguistics: EMNLP 2024},
  pages={7346--7356},
  year={2024}
}

@article{meincke2025prompting,
  title={Prompting science report 2: The decreasing value of chain of thought in prompting},
  author={Meincke, Lennart and Mollick, Ethan and Mollick, Lilach and Shapiro, Dan},
  journal={arXiv preprint arXiv:2506.07142},
  year={2025}
}

@article{kojima2022large,
  title={Large language models are zero-shot reasoners},
  author={Kojima, Takeshi and Gu, Shixiang Shane and Reid, Machel and Matsuo, Yutaka and Iwasawa, Yusuke},
  journal={Advances in neural information processing systems},
  volume={35},
  pages={22199--22213},
  year={2022}
}

@article{hendrycks2021measuring,
  title={Measuring mathematical problem solving with the math dataset},
  author={Hendrycks, Dan and Burns, Collin and Kadavath, Saurav and Arora, Akul and Basart, Steven and Tang, Eric and Song, Dawn and Steinhardt, Jacob},
  journal={arXiv preprint arXiv:2103.03874},
  year={2021}
}

@article{yang2024chain,
  title={Chain-of-thought in neural code generation: From and for lightweight language models},
  author={Yang, Guang and Zhou, Yu and Chen, Xiang and Zhang, Xiangyu and Zhuo, Terry Yue and Chen, Taolue},
  journal={IEEE Transactions on Software Engineering},
  volume={50},
  number={9},
  pages={2437--2457},
  year={2024},
  publisher={IEEE}
}

@article{an2025amo,
  title={Amo-bench: Large language models still struggle in high school math competitions},
  author={An, Shengnan and Cai, Xunliang and Cao, Xuezhi and Li, Xiaoyu and Lin, Yehao and Liu, Junlin and Lv, Xinxuan and Ma, Dan and Wang, Xuanlin and Wang, Ziwen and others},
  journal={arXiv preprint arXiv:2510.26768},
  year={2025}
}

@article{cobbe2021gsm8k,
  title={Training Verifiers to Solve Math Word Problems},
  author={Cobbe, Karl and Kosaraju, Vineet and Bavarian, Mohammad and Chen, Mark and Jun, Heewoo and Kaiser, Lukasz and Tworek, Jerry and Wang, Peter and Chen, Xi and Perez, Ethan and others},
  journal={arXiv preprint arXiv:2110.14168},
  year={2021}
}

@article{ke2025survey,
  title={A survey of frontiers in llm reasoning: Inference scaling, learning to reason, and agentic systems},
  author={Ke, Zixuan and Jiao, Fangkai and Ming, Yifei and Nguyen, Xuan-Phi and Xu, Austin and Long, Do Xuan and Li, Minzhi and Qin, Chengwei and Wang, Peifeng and Savarese, Silvio and others},
  journal={arXiv preprint arXiv:2504.09037},
  year={2025}
}

@article{zhang2024llm,
  title={Llm as a mastermind: A survey of strategic reasoning with large language models},
  author={Zhang, Yadong and Mao, Shaoguang and Ge, Tao and Wang, Xun and de Wynter, Adrian and Xia, Yan and Wu, Wenshan and Song, Ting and Lan, Man and Wei, Furu},
  journal={arXiv preprint arXiv:2404.01230},
  year={2024}
}

@inproceedings{tang2025llm,
  title={Llm integration in extended reality: A comprehensive review of current trends, challenges, and future perspectives},
  author={Tang, Yiliu and Situ, Jason and Cui, Andrea Yaoyun and Wu, Mengke and Huang, Yun},
  booktitle={Proceedings of the 2025 CHI Conference on Human Factors in Computing Systems},
  pages={1--24},
  year={2025}
}

@article{wu2025role,
  title={On the Role of Temperature Sampling in Test-Time Scaling},
  author={Wu, Yuheng and Mirhoseini, Azalia and Tambe, Thierry},
  journal={arXiv preprint arXiv:2510.02611},
  year={2025}
}

@article{zelikman2022star,
  title={Star: Bootstrapping reasoning with reasoning},
  author={Zelikman, Eric and Wu, Yuhuai and Mu, Jesse and Goodman, Noah},
  journal={Advances in Neural Information Processing Systems},
  volume={35},
  pages={15476--15488},
  year={2022}
}

@article{wang2025sampling,
  title={Sampling-efficient test-time scaling: Self-estimating the best-of-n sampling in early decoding},
  author={Wang, Yiming and Zhang, Pei and Huang, Siyuan and Yang, Baosong and Zhang, Zhuosheng and Huang, Fei and Wang, Rui},
  journal={arXiv preprint arXiv:2503.01422},
  year={2025}
}

@misc{xai2025grok41,
  author       = {{xAI}},
  title        = {Grok 4.1 Model Card},
  howpublished = {\url{https://data.x.ai/2025-11-17-grok-4-1model-card.pdf}},
  year         = {2025},
  month        = {Nov},
  note         = {Accessed: 2025-12-12}
}

@inproceedings{liu2024automatic,
  title={Automatic generation of model and data cards: A step towards responsible AI},
  author={Liu, Jiarui and Li, Wenkai and Jin, Zhijing and Diab, Mona},
  booktitle={Proceedings of the 2024 Conference of the North American Chapter of the Association for Computational Linguistics: Human Language Technologies (Volume 1: Long Papers)},
  pages={1975--1997},
  year={2024}
}

@article{du2025optimizing,
  title={Optimizing temperature for language models with multi-sample inference},
  author={Du, Weihua and Yang, Yiming and Welleck, Sean},
  journal={arXiv preprint arXiv:2502.05234},
  year={2025}
}
}

\end{document}